\documentclass[runningheads]{llncs}
\PassOptionsToPackage{dvipsnames,table}{xcolor}

\usepackage{eccvabbrv}

\usepackage{graphicx}
\usepackage{booktabs}
\usepackage{multirow}
\usepackage{wrapfig}
\usepackage{algorithm}
\usepackage{algorithmic} 
\usepackage[table]{xcolor}
\usepackage{tcolorbox}
\tcbuselibrary{breakable}
\usepackage{fvextra}

\usepackage[accsupp]{axessibility}  

\usepackage{hyperref}

\usepackage{orcidlink}
\usepackage{pifont}
\newcommand{\cmark}{\ding{51}}
\newcommand{\xmark}{\ding{55}}

\begin{document}

\title{CrossView: Can Vision-Language Models Reason Across Cameras?} 

\titlerunning{CrossView}

\newcommand{\samethanks}[1][\value{footnote}]{\footnotemark[#1]}
\author{
Sahil Shah\thanks{Equal contribution.}\inst{1}\orcidlink{0009-0009-3640-2339}\and
S P Sharan\samethanks\inst{1}\orcidlink{0000-0002-6298-6464}\and
Harsh Goel\samethanks\inst{1}\orcidlink{0009-0006-9873-9584}\and
Manvik Pasula\inst{2}\orcidlink{0009-0007-3390-5198}\and
Adithya Hebbalae\inst{1}\orcidlink{0009-0003-7454-4369}\and
Minkyu Choi\inst{1}\orcidlink{0009-0007-6557-8865}\and
Sandeep P. Chinchali\inst{1}\orcidlink{0000-0002-0601-3633}}
\authorrunning{S.~Shah et al.}

\institute{
The University of Texas at Austin, Austin TX, USA \and
Independent Researcher, USA
}

\maketitle

\begin{abstract}
    Video understanding benchmarks have long centered on single\nobreakdash-camera settings, where modern multi-modal language models achieve strong performance across image and video tasks. Yet, the real world runs on multi-camera networks: autonomous vehicles, security systems, and robots all gather data across many simultaneous views. We argue that this is not simply ``more'' of the single-camera problem; it is fundamentally different. Multi-camera reasoning requires handling context that scales with the number of views, resolving occlusions visible from only a subset of cameras, judging which views matter, and integrating evidence across perspectives that may overlap or diverge. Current models struggle with exactly these challenges, yet no benchmark systematically targets them. We introduce \texttt{CrossView}, a multi-camera video question-answering benchmark spanning autonomous driving, security surveillance, egocentric/exocentric video, and robotics. Evaluation of proprietary models, such as GPT-5.2, and open-source models, like Qwen3-VL, reveals consistently low accuracy, with open-source models trailing by a wide margin. Performance scales strongly with a model's ability to jointly process multiple viewpoints, positioning \texttt{CrossView} as a rigorous benchmark for multi-camera video. We open-source our code and dataset at \url{https://utaustin-swarmlab.github.io/CrossView}.
\end{abstract}

\section{Introduction}
The ``single-camera assumption'' has long dominated the landscape of computer vision. From image classification to recent Large Vision-Language Models (LVLMs), benchmarks have primarily evaluated the ability to reason over a single camera. Most state-of-the-art (SOTA) models now excel at this, reaching near-human performance on visual benchmarks such as VQAv2 \cite{goyal2017vqav2}, and video benchmarks like Video-MME \cite{fu2024videomme} and LongVideoBench \cite{wu2024longvideobench}.

However, real-world systems rarely rely on a single perspective. In autonomous driving (AD), robotics, and wide-area surveillance, perception is inherently multi-camera: a self-driving car fuses surround-view cameras to navigate intersections, a robotic arm coordinates wrist-mounted and overhead views to manipulate objects, and security networks track subjects across disjoint fields of view. Despite this, current multi-modal language models are still evaluated almost exclusively on single-camera inputs. 

The gap between single- and multi-camera video understanding is not merely one of data volume. Multi-camera reasoning introduces two fundamental \textit{peculiarities} that current models find uniquely challenging: \ding{202} Context Scaling, where processing $N$ simultaneous streams expands the required context window by an order of magnitude and often exceeds the effective window of long-context models, and \ding{203} Cross-View Spatial Reasoning, where models must go beyond single-camera temporal and visual reasoning to perform spatio-temporal stitching (\textit{i.e.}, synthesizing discrete viewpoints into a coherent scene). This entails distinguishing overlapping \textit{vs.} non-overlapping fields of view, selecting the most informative camera for a specific sub-task (camera identification), aggregating evidence across cameras (\textit{e.g.}, counting unique objects), and reasoning about events across disjoint streams.

\begin{figure}[t]
    \centering
    \includegraphics[width=0.95\linewidth]{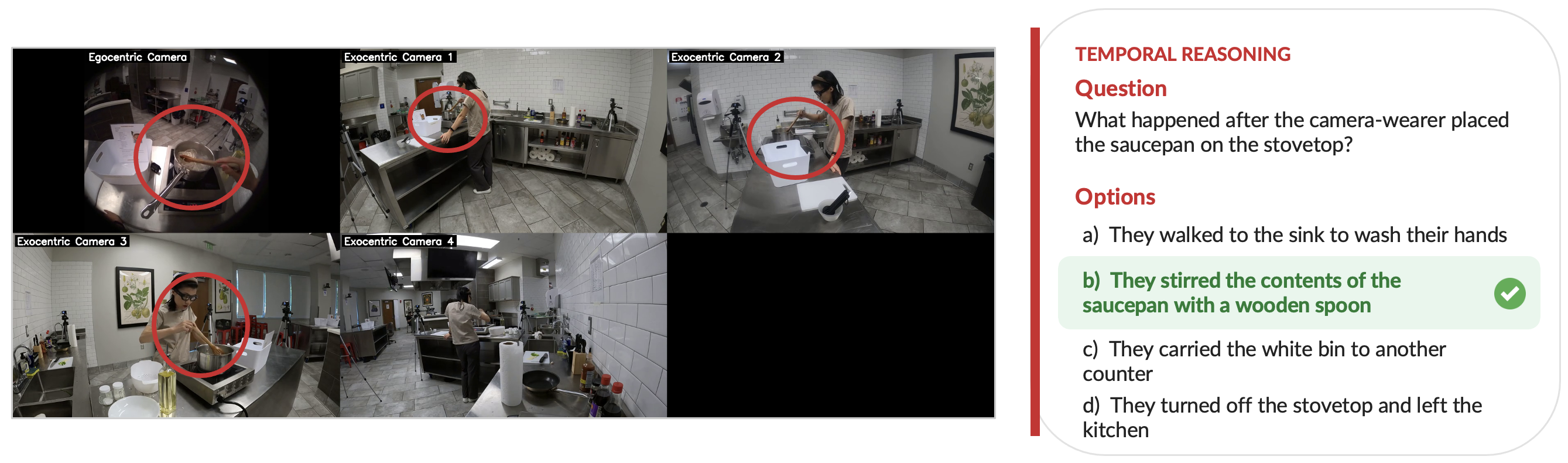}
    \caption{\textbf{\texttt{CrossView} data example illustrating multi-view reasoning}. Multiple synchronized camera streams observe the \textit{same} scene from \textit{different} viewpoints (egocentric and exocentric). Answering the question requires integrating information across views (e.g., to identify the event that occurs after the saucepan is placed on the stovetop). \texttt{CrossView} evaluates such cross-view reasoning tasks across diverse domains, including robotics, surveillance, autonomous driving, and ego–exo interaction.}
    \label{fig:teaser}
\end{figure}

Existing datasets supply raw multi-camera data, but they primarily only address narrow facets of cross-view reasoning. nuScenes-QA \cite{qian2024nuscenesqa,caesar2020nuscenes} targets only perception, motivating methods that merge views into a unified Bird's-Eye-View (BEV) representation \cite{ding2024holistic}, while Ego-Exo4D \cite{grauman2024egoexo4d} is restricted to activity understanding. There is a distinct lack of benchmarks that force a model to reason spatio-temporally across raw, simultaneous, and heterogeneous camera feeds. 

We address this with \texttt{CrossView}, a comprehensive multi-camera video question answering (VQA) benchmark designed to stress-test the cross-view reasoning capabilities of modern VLMs. \texttt{CrossView} comprises 6,000 questions across four domains (Autonomous Driving, Surveillance, Egocentric-Exocentric interaction, and Robotics) with tasks such as camera identification, cross-view object counting, and spatio-temporal event ordering that cannot be solved from any single camera in isolation, a property we verify empirically by restricting inputs to a single view (Table~\ref{tab:best_camera}).

Our contributions can be summarized as follows:
\begin{enumerate}
    \item Novel Multi-Camera Benchmark: We introduce \texttt{CrossView}, the first VQA dataset specifically curated to evaluate joint reasoning over 2 to 8 simultaneous camera streams across four real-world domains.
    \item Targeted Multi-View Reasoning Tasks: We design natural-language questions with five distinct categories (Temporal, Event Ordering, Spatial, Counting, and Summarization) alongside a ``Camera-ID'' task that requires models to understand spatially across multiple viewpoints and temporally across the video to answer questions.
    \item Comprehensive Model Evaluation: We provide an extensive study of both proprietary (GPT-5.2) and open-source (Qwen, InternVL, and Gemma families) models, revealing a significant ``multi-camera gap'' where performance degrades in multi-camera settings.
    \item Multi-view Visual Aggregation: We investigate how different architectural choices of stacking frames from multiple videos as visual input impact a model's ability to maintain spatial-temporal understanding across views.
\end{enumerate}

\section{Related Works}
\subsection{Visual Question Answering Benchmarks and Models}

\paragraph{Image and Video QA.}
Visual question answering began at the image level: VQAv2 \cite{goyal2017vqav2} paired open-ended questions with natural images and established accuracy as the standard metric. GQA~\cite{hudson2019gqa} added compositional scene-graph-grounded questions for spatial and relational reasoning, and OK-VQA \cite{marino2019okvqa} and A-OKVQA \cite{schwenk2022aokvqa} emphasized questions requiring external knowledge. Video QA followed with early benchmarks such as TGIF-QA \cite{jang2017tgif} and ActivityNet-QA \cite{yu2019activitynetqa} targeting spatio-temporal reasoning over short video clips. Recent long-video methods \cite{ye2025re,shah2026neus,wang2025videotree} and benchmarks have also emerged: LongVideoBench \cite{wu2024longvideobench} assesses interleaved video-language understanding, Video-MME \cite{fu2024videomme} spans short to hour-long videos, MLVU \cite{zhou2024mlvu} focuses on multi-task long video understanding, EgoSchema \cite{mangalam2024egoschema} examines egocentric temporal reasoning, and MVBench \cite{li2024mvbench} introduces evaluation for temporal perception. Despite this progress, all primarily still rely on a \emph{single} camera stream, leaving cross-view evidence aggregation, view selection, and context scaling with the number of cameras entirely unexplored.

\paragraph{Driving-domain QA.}
Driving benchmarks work with multi-camera data but still lack true multi-camera reasoning. NuScenes-QA~\cite{qian2024nuscenesqa} builds QA pairs on nuScenes from 3D scene graphs but stays tied to spatial perception, while NuPlanQA~\cite{luo2025nuplanqa} and OmniDrive~\cite{wang2024omnidrive} add planning-oriented temporal reasoning that remains ego-centric. The TLV dataset~\cite{choi2024towards} further focuses on search problems in autonomous driving scenarios. Each moves evaluation closer to real-world driving, however they still abstract away core multi-camera challenges such as allocentric spatio-temporal reasoning. \texttt{CrossView} directly tests this by forcing integration across concurrent video streams.

\paragraph{Vision-language models.} 
VLMs are advancing in step with these benchmarks. Proprietary models like GPT-4o~\cite{openai2024gpt4o} and Gemini~\cite{team2024gemini} support long visual contexts and excel on single-camera inputs. Open-source models such as Qwen-VL~\cite{bai2025qwen2,bai2025qwen3vl} and InternVL~\cite{chen2024internvl,chen2025internvl25} handle variable-length multi-image and video inputs, while Gemma~\cite{team2025gemma3} offers lightweight options. All, however, are designed and evaluated for single-view inputs, leaving multi-feed reasoning unevaluated. \texttt{CrossView} addresses this gap.

\subsection{Multi-camera Systems and Multi-View Understanding}

Multi-camera data is ubiquitous in deployed systems, and a rich body of work addresses perception and understanding over it. We survey the key areas below.

\paragraph{Multi-camera perception and tracking.}
Multi-target multi-camera (MTMC) tracking is a central problem, with graph-based methods~\cite{ristani2018features,wen2017multi} jointly optimizing detection and association across views, and benchmarks such as CityFlow~\cite{tang2019cityflow} and WILDTRACK~\cite{chavdarova2018wildtrack} establishing cross-camera video understanding performance in urban and surveillance settings. Approaches such as MVDet~\cite{hou2021multiview} improve cross-view understanding on these datasets via convolutional neural networks. In autonomous driving, nuScenes~\cite{caesar2020nuscenes} provides synchronized six-camera feeds used for 3D detection~\cite{wang2022detr3d,liu2023petrv2}, BEV segmentation~\cite{li2023bevformer,philion2020lss}, and tracking~\cite{chiu2021probabilistic}. Cross-camera person re-identification reasons across views to match identities, driven by large-scale datasets such as Market-1501~\cite{zheng2015market}, MSMT17~\cite{wei2018msmt17}, CUHK03~\cite{li2014deepreid}, and MARS~\cite{zheng2016mars} and diverse methods ~\cite{hermans2017triplet,luo2019bag}. The MEVA dataset~\cite{corona2021meva}, used in \texttt{CrossView}, was originally designed for activity detection in large surveillance networks. These methods perform genuine cross-view reasoning, but they target low-level perception outputs, such as bounding boxes, trajectories, and identity embeddings, rather than the language-grounded semantic understanding, event ordering, and spatial summarization \texttt{CrossView} requires.
\paragraph{Multi-view 3D understanding.}
A related line of work uses multiple viewpoints for 3D scene understanding. Multi-view stereo~\cite{yao2018mvsnet,gu2020cascade} reconstructs dense geometry from calibrated images, bird's-eye-view methods~\cite{li2023bevformer,philion2020lss} project surround-view features into a unified spatial representation for driving. Neural radiance fields~\cite{mildenhall2020nerf}, and 3D Gaussian splatting~\cite{kerbl2023gaussian} synthesize novel views from multi-view inputs, with recent work~\cite{hong20233dllm,zhu2024llava3d} coupling 3D representations to language models for spatial QA. All, however, depend on a reconstructed or unified 3D representation rather than reasoning directly over raw multi-camera inputs.

\paragraph{Egocentric-exocentric understanding.}
Ego-Exo4D~\cite{grauman2024egoexo4d} provides synchronized egocentric and exocentric views of skilled activities with keystep, proficiency, and language annotations, building on Ego4D~\cite{grauman2022ego4d}, which established large-scale egocentric benchmarks for episodic memory, forecasting, and social interaction. Recent work increasingly links the two perspectives: He~\emph{et al.}~\cite{he2026bridging} survey ego-exo understanding, Exo2Ego~\cite{zhang2026exo2ego} examines knowledge transfer across views, Ego2ExoVLM~\cite{reilly2025my} tests the recognition of egocentric content from exocentric views, and EgoExoBench~\cite{he2026egoexobench} evaluates VLMs on ego-exo relationships. PDB-Eval~\cite{wu2025pdb} measures a driver's behavior in an ego-exo setting, and LangView~\cite{majumder2025viewpoint} studies viewpoint selection on Ego-Exo4D \cite{grauman2024egoexo4d} and LEMMA \cite{jia2020lemma}, directly paralleling the camera-identification task in \texttt{CrossView}. These existing benchmarks focus on viewpoint correspondence or selection in isolation; none require \emph{joint} integration of evidence across all cameras for counting, spatial reasoning, temporal ordering, and summarization questions---capabilities \texttt{CrossView} targets.

\begin{table}[t]
\centering
\caption{Comparison of \texttt{CrossView} with existing VQA benchmarks. \textbf{\#Cam}: number of simultaneous camera views. \textbf{\#Q}: number of questions. \textbf{Cross-View}: whether questions require integrating evidence across camera perspectives. \textbf{Cam-ID}: whether the benchmark includes camera identification/selection questions. \textbf{Multi-domain}: whether the benchmark spans multiple domains. \textbf{Questions}: Temp (temporal), Evt (event ordering), Spat (spatial), Count (counting), Summ (summarization).}
\label{tab:benchmark_comparison}
\resizebox{\textwidth}{!}{%
\begin{tabular}{lcccccccccccc}
\toprule
\textbf{Benchmark} & \textbf{\#Cam} & \textbf{\#Q} & \textbf{Duration} & \textbf{Domains} & \textbf{Cross-View} & \textbf{Cam-ID} & \textbf{Multi-domain} & \textbf{Temp} & \textbf{Evt} & \textbf{Spat} & \textbf{Count} & \textbf{Summ} \\
\midrule
\multicolumn{13}{l}{\textit{Image \& Short-Video QA}} \\
VQAv2~\cite{goyal2017vqav2} & 1 & 1.4M & Image & Natural & \xmark & \xmark & \xmark & \xmark & \xmark & \cmark & \cmark & \xmark \\
GQA~\cite{hudson2019gqa} & 1 & 22M & Image & Natural & \xmark & \xmark & \xmark & \xmark & \xmark & \cmark & \xmark & \xmark \\
GazeVQA~\cite{chen2023gazevqa} & 3 & 25K & $\sim$2.5 min & Gaze (ego-exo) & \cmark & \xmark & \xmark & \xmark & \xmark & \cmark & \xmark & \xmark \\
TGIF-QA~\cite{jang2017tgif} & 1 & 165K & $<$10\,s & GIFs & \xmark & \xmark & \xmark & \cmark & \xmark & \xmark & \cmark & \xmark \\
ActivityNet-QA~\cite{yu2019activitynetqa} & 1 & 58K & $\sim$3 min & Activities & \xmark & \xmark & \xmark & \cmark & \xmark & \cmark & \cmark & \xmark \\
\midrule
\multicolumn{13}{l}{\textit{General Video QA}} \\
EgoSchema~\cite{mangalam2024egoschema} & 1 & 5K & $\sim$3 min & Egocentric & \xmark & \xmark & \xmark & \cmark & \cmark & \xmark & \xmark & \cmark \\
MVBench~\cite{li2024mvbench} & 1 & 4K & $\sim$16\,s & Multi-source & \xmark & \xmark & \cmark & \cmark & \cmark & \cmark & \cmark & \xmark \\
LongVideoBench~\cite{wu2024longvideobench} & 1 & 6.7K & $\sim$8 min & Multi-source & \xmark & \xmark & \cmark & \cmark & \cmark & \cmark & \xmark & \xmark \\
VideoMME~\cite{fu2024videomme} & 1 & 2.7K & $\sim$17 min & Multi-source & \xmark & \xmark & \cmark & \cmark & \cmark & \cmark & \cmark & \xmark \\
MLVU~\cite{zhou2024mlvu} & 1 & 3.1K & $\sim$15.5 min & Multi-source & \xmark & \xmark & \cmark & \cmark & \cmark & \xmark & \cmark & \cmark \\
\midrule
\multicolumn{13}{l}{\textit{Driving-Domain QA}} \\
NuScenes-QA~\cite{qian2024nuscenesqa} & 6$^\dagger$ & 460K & Seconds & Driving & \xmark & \xmark & \xmark & \cmark & \xmark & \cmark & \cmark & \xmark \\
NuPlanQA~\cite{luo2025nuplanqa} & 8 & 1M & Seconds & Driving & \xmark & \xmark & \xmark & \cmark & \xmark & \cmark & \xmark & \xmark \\
OmniDrive~\cite{wang2024omnidrive} & 6$^\dagger$ & 300K & Seconds & Driving & \xmark & \xmark & \xmark & \cmark & \xmark & \cmark & \xmark & \xmark \\
\midrule
\multicolumn{13}{l}{\textit{Ego-Exo Benchmarks}} \\
EgoExoBench~\cite{he2026egoexobench} & 2--5 & 7.3K & Minutes & Ego-exo & Partial & \xmark & \xmark & \cmark & \cmark & \cmark & \xmark & \xmark \\
LangView~\cite{majumder2025viewpoint} & 2--5 & -- & Minutes & Ego-exo & Partial & \cmark & \xmark & \xmark & \xmark & \xmark & \xmark & \xmark \\
PDB-Eval~\cite{wu2025pdb} & 2 & 44.9K & Minutes & Ego-exo (driving) & Partial & \xmark & \xmark & \cmark & \xmark & \xmark & \xmark & \xmark \\
\midrule
\texttt{CrossView} \textbf{(Ours)} & \textbf{2--8} & \textbf{6K} & \textbf{Sec--min} & \textbf{AD, Surv., Ego, Robot.} & \cmark & \cmark & \cmark & \cmark & \cmark & \cmark & \cmark & \cmark \\
\bottomrule
\multicolumn{13}{l}{\footnotesize $^\dagger$Data sourced from 6 cameras, but questions are grounded to individual views and do not require cross-view reasoning.} \\
\end{tabular}%
}
\end{table}

\paragraph{Robotic multi-camera systems.}
Multi-camera setups are standard in robotic manipulation and navigation. Datasets such as AgiBot~\cite{bu2025agibot} provide multi-view recordings from wrist-mounted and external cameras. Prior robotic video benchmarks~\cite{zitkovich2023rt,driess2023palme} target action prediction and planning from single or paired views rather than language-grounded reasoning across simultaneous feeds. \texttt{CrossView} draws robotic scenarios from AgiBot to test whether VLMs can reason about manipulation when evidence is distributed across cameras.

\subsection{Benchmark Comparison}

Table~\ref{tab:benchmark_comparison} positions \texttt{CrossView} against existing VQA benchmarks. While prior work thoroughly covers single-camera settings across varied durations and question types, none systematically evaluate multi-camera reasoning. \texttt{CrossView} is the first to combine simultaneous camera views, diverse real-world domains, and question categories designed specifically to probe cross-view spatio-temporal reasoning.

\nocite{goel2026incentivizing, shah2025challenge, liang2026neus}

\section{Dataset Construction}
We first construct a dataset via a two-stage pipeline: we build a Spatio-temporal Scene Graph (STSG) from existing robotics, surveillance, and autonomous driving datasets, then apply a programmatic question-generation engine. The STSG is a structured representation consolidating semantic and event-centric metadata into a unified representation which the question-generation engine queries to synthesize complex questions without risk of model hallucination. We first programmatically identify grounding events/objects, target events, correct ground-truth answers, and plausible distractors from the scene graphs, then pass this metadata to GPT-5.2 to phrase natural-language questions, ensuring semantic correctness without human verification.

\subsection{Spatio-temporal Scene Graph (STSG) Construction}

\begin{algorithm}[t]
\caption{Spatio-temporal Scene Graph (STSG) Construction}
\label{alg:stsg-construction}
\begin{algorithmic}[1]
\REQUIRE Multi-modal datasets $\mathcal{D} \in {\text{nuScenes, Ego-Exo4D, AgiBot, MEVA}}$, video timestamps $\mathcal{T}$
\ENSURE Spatio-temporal Scene Graph $\mathcal{G}$

\STATE \textbf{Phase 1: Multi-modal perception and metadata consolidation}
\FOR{each scene in $\mathcal{D}$}
\STATE Extract object trajectories and spatial coordinates
\IF{dataset is nuScenes}
\STATE Refine 3D localization using LiDAR point cloud data
\ELSE
\STATE Utilize provided 3D or bounding-box annotations
\ENDIF
\STATE Generate object-centric activity labels and descriptions using VLM-based annotation for nuScenes/AgiBot or native metadata for Ego-Exo4D/MEVA
\STATE Generate timestamp-level scene captions using GPT-5.2 from annotated objects and activities
\STATE Compute event intervals $[t_{start}, t_{end}]$ by grouping consecutive timestamps with consistent object activities
\ENDFOR

\STATE \textbf{Phase 2: Spatial and temporal graph construction}
\STATE Initialize $\mathcal{G}=\emptyset$
\FOR{each timestamp $t \in \mathcal{T}$}
\STATE Compute pairwise spatial predicates $R_t$ between object pairs, e.g., \textit{behind}, \textit{near}, \textit{left}, and \textit{right}
\STATE Calculate orientation deltas and relative distances between all object pairs
\STATE Obtain scene graph snapshot $\mathcal{G}_t = {\mathcal{N}_t, \mathcal{E}_t}$, where $\mathcal{N}_t$ contains object nodes with activity labels, descriptions, and event intervals, and $\mathcal{E}_t$ contains spatial relationship edges between objects
\STATE $\mathcal{G} = \mathcal{G} \cup \mathcal{G}_t$
\ENDFOR
\RETURN $\mathcal{G}$
\end{algorithmic}
\end{algorithm}

The STSG construction in Algorithm \ref{alg:stsg-construction} begins by consolidating metadata from Ego-Exo4D, nuScenes, AgiBot, and MEVA. We use the spatial coordinates of annotated objects across these datasets [Line 3]: for nuScenes we incorporate LiDAR point clouds for 3D localization [Line 5], while for the others we rely on the provided object-centric 3D or bounding-box annotations [Line 7]. To add semantic information (\textit{i.e.}, object activities and descriptions), we use InternVL-3.5 38B to annotate object-centric activities for datasets lacking native labels [Line 9]: for nuScenes and AgiBot we first associate each object with its bounding box before querying InternVL for corresponding activity descriptions. In contrast, Ego-Exo4D and MEVA supply action and event labels natively. We additionally parse per-frame annotated objects and activities with GPT-5.2 to obtain a scene-level caption per timestamp.

Once objects are localized and their activities identified, we compute pairwise spatial relationships between all entities in the scene across camera views. These relationships include directional predicates such as \textit{behind}, \textit{near}, \textit{left}, and \textit{right}, plus orientation deltas and relative distances [Lines 13--14]. We then form event intervals $[t_{start}, t_{end}]$ by grouping consecutive timestamps over which a given object maintains the same activity [Line 15]; we denote the start and end timestamp of an event $E$'s interval by $E$.\texttt{start} and $E$.\texttt{end} respectively. These intervals are derived from the object activity labels obtained in Line 9.

Finally, for each timestamp $t$, we generate a spatio-temporal graph snapshot ($\mathcal{G}_t = {\mathcal{N}_t, \mathcal{E}_t}$), where $\mathcal{N}_t$ denotes object nodes containing activity labels, descriptions, and associated event intervals, and $\mathcal{E}_t$ denotes edges encoding spatial relationships between objects. These graph snapshots are appended chronologically to yield the final STSG $\mathcal{G}$ [Line 16].

\subsection{Programmatic Question Construction}

Our question construction follows a Grounding-Target architecture that programmatically queries the STSG. For each question, we identify grounding event(s) or object(s) that serve as anchors, select the corresponding target event or object to be queried, and sample negative candidates that act as distractors. This structured information, consisting of the grounding anchors, target answer, and distractors, is passed to GPT-5.2 which generates a coherent natural-language question. We provide the QA generation prompts in the Appendix.

\subsubsection{Temporal Reasoning.}

We use the temporally consolidated events in the STSG to construct temporal and spatio-temporal questions. For temporal questions, we sample a grounding event ($E_g$) and a target event ($E_t$), and assign one of four temporal relations: \textit{Before}, \textit{After}, \textit{During}, or \textit{In-between}. \textit{Before} holds when ($E_t$.\texttt{end} $<$ $E_g$.\texttt{start}), while \textit{After} holds when ($E_t$.\texttt{start} $>$ $E_g$.\texttt{end}). \textit{During} holds when the temporal overlap between the two events exceeds $50\%$ of the shorter event's duration. \textit{In-between} is defined using two non-overlapping grounding events, ($E_{g1}$) and ($E_{g2}$), where the target event lies entirely within the temporal gap between them. For example, this relation can produce questions such as: ``What happens in between a blue car parked by the street and the pedestrian with a blue shirt walking across the crosswalk?''

We extend this to spatio-temporal cross-referencing via two categories. In Category I (Spatial-to-Temporal), the grounding anchor is a spatial state, defined by a directional predicate between two objects at a given frame, and the target is an adjacent or intervening temporal activity. For example, a question could be: ``When the pedestrian wearing the white shirt approaches and is to the left of the black parked car, what happens immediately after?'' In Category II (Temporal-to-Spatial), the grounding anchor is a temporal boundary, and the target is a spatial snapshot, an intervening relation between two events, or a comparative change in object relations. For example, this category can generate questions such as: ``What happens in between a woman cycling on the white bike taking a left turn and the yellow car stopping at the intersection?'' with the answer: ``A delivery truck stops behind the yellow car.''

Finally, we convert the sampled tuples into natural-language questions with GPT-5.2. The prompt enforces grounded descriptions by referencing object appearance and activity rather than generic class labels. To increase difficulty, we sample three distractor types: spatial distractors (wrong direction at the correct time), temporal distractors (correct event at the wrong time), and existential distractors (objects absent from the scene). Each sample is stored as a JSON object containing the question, distractors, and reasoning.

\begin{figure}[t]
    \centering
    \includegraphics[width=0.9\linewidth]{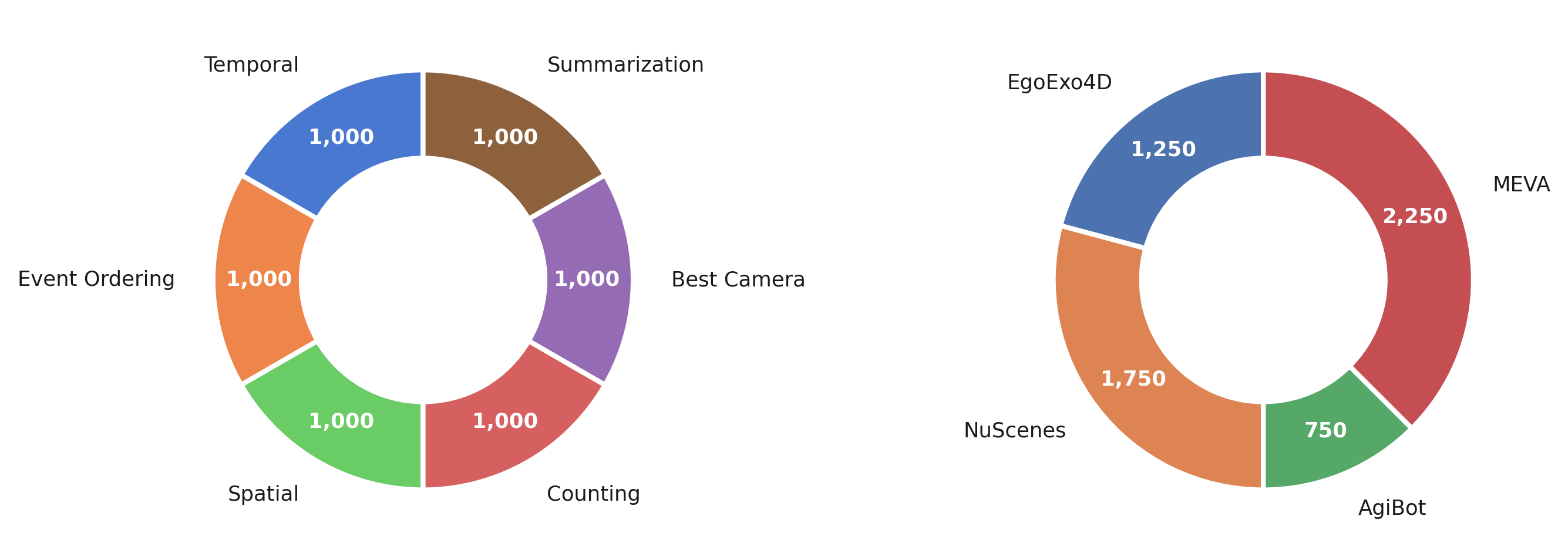}%
    \caption{Dataset question distribution. \textbf{Left: } Questions are distributed among six categories, each contributing 1,000 questions. \textbf{Right: } Distributions across the four source datasets. MEVA and nuScenes contribute the majority of the questions  (2,250 and 1,750 respectively), while Ego-Exo4D and AgiBot provide 1,250
   and 750 questions.}
    \label{fig:dataset_distribution}
\end{figure}

\subsubsection{Spatial Question.}

Spatial questions test the model's understanding of 3D environments from multiple viewpoints. For egocentric tasks, the engine extracts directional predicates of objects relative to the ego-vehicle. For frame-of-reference tasks, it selects a non-ego reference object, retrieves its orientation quaternion, and computes target positions in that object's local coordinate frame. Multi-hop spatial queries are created by choosing a reference node and at least two target objects at clearly different metric distances, enabling comparative-proximity questions grounded in spatial metadata rather than 2D pixel distance. For example: ``Where is the traffic pole relative to the parked blue car that is parked directly adjacent to the main road?''

\subsubsection{Global Scene Summarization.}

Summarization evaluates the model's ability to comprehensively and holistically understand a long video. Rather than querying a single timestamp, the engine aggregates the entire STSG into a dense temporal narrative by sampling frame-level captions and significant object-centric events at regular intervals. The resulting timeline gives a compressed yet representative overview of the scene's dynamics, which prompts the model to produce a comprehensive summary accounting for the ego-actor's primary interactions and the evolution of the environment across all camera perspectives.

\begin{figure}[t]
    \centering
    \includegraphics[width=\linewidth]{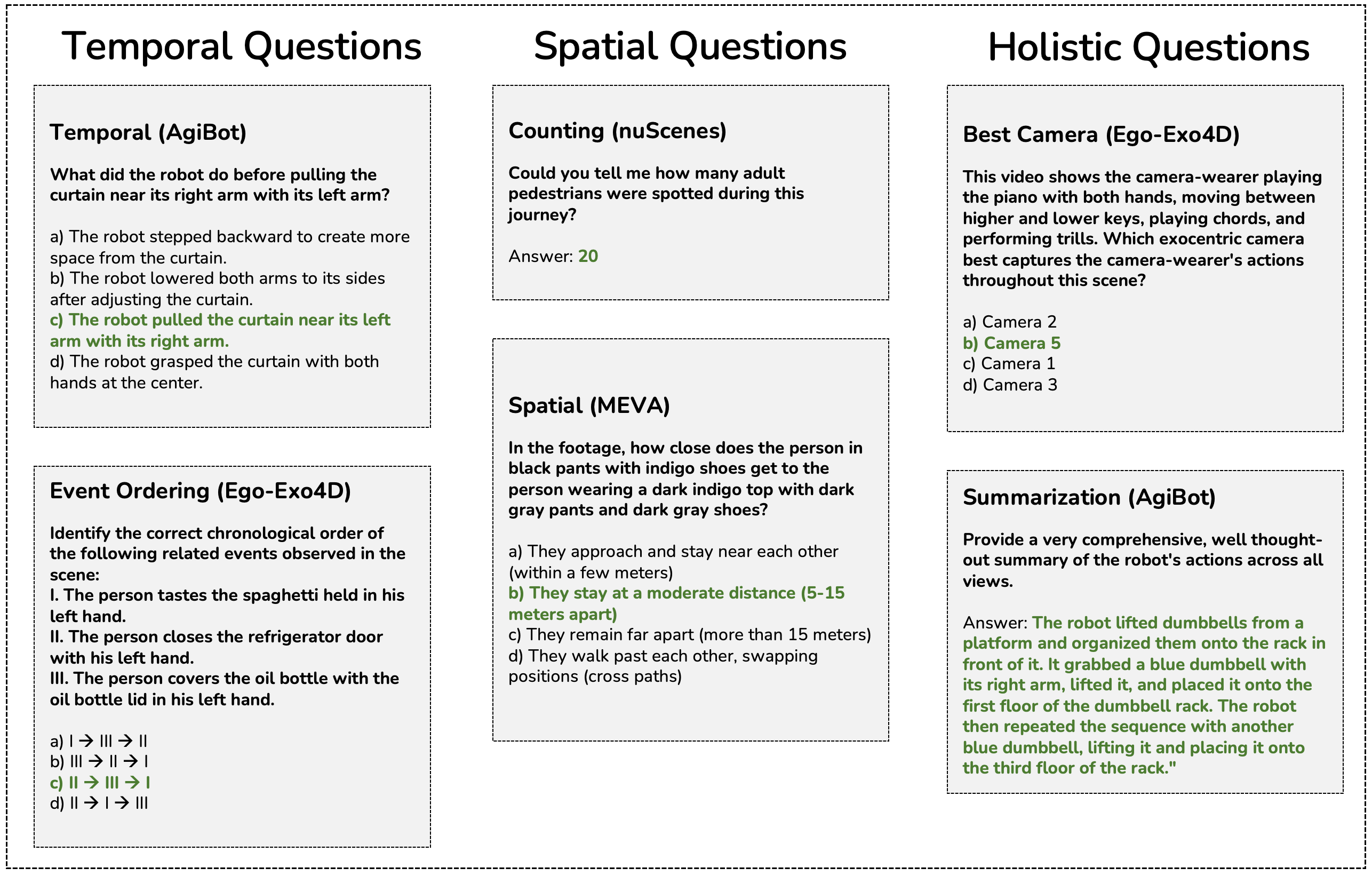}
    \caption{\textbf{Question categories in \texttt{CrossView}.} The benchmark includes six types of multi-camera reasoning tasks grouped into three categories: temporal reasoning (temporal queries and event ordering), spatial reasoning (counting and spatial relationships), and holistic understanding (best-camera selection and scene summarization). Each question requires integrating information across multiple synchronized camera views and spans diverse domains including robotics (AgiBot), autonomous driving (nuScenes), surveillance (MEVA), and egocentric–exocentric interaction (Ego-Exo4D).}
    \label{fig:questions}
\end{figure}

\subsubsection{Optimal Viewpoint Selection (Best Camera).}

The best camera category evaluates whether a model can identify the most informative perspective for a given event. The engine tracks each targeted object or activity's visibility metadata across all camera sensors and identifies the ``optimal'' camera view using the most informative perspective available in the STSG. Specifically, this is the natively annotated best exocentric view in Ego-Exo4D or the camera that most persistently observes the activity in MEVA. The resulting question presents an activity and asks the model which camera provided the most persistent or clear visual evidence, effectively testing spatial-semantic data indexing. 

\subsubsection{Object Instance Counting.}

Counting questions test the model's ability to distinguish and track individual instances over time. To do this, the engine counts unique identifiers (UIDs) across the full temporal span of the STSG, restricted to selected semantic classes. Aggregating UIDs rather than per-frame detections keeps the ground truth accurate under occlusion, reappearance, or movement between camera views, and requires persistent identity tracking across the entire video. An example of this is: ``How many pedestrians are located near the red building directly adjacent to the main road?''

\subsubsection{Chronological Event Ordering.}

Event ordering tests the model's ability to reconstruct a scene's temporal flow. The engine scans the STSG for a chain of three to five distinct, diverse events and, via a buffer-frame heuristic, selects events forming a clear chronological sequence with minimal overlap. Once shuffled and presented to the model, they must be restored to their original chronological order using semantic and temporal cues from the video, evaluating how well the model links distinct object-centric actions into a coherent timeline. An example is: ``Read the events below and organize them as they appear in the multi-camera video stream: 1. The delivery truck appears and stops behind the yellow car, 2. The yellow car stops at the intersection, 3. A pedestrian crosses the road, 4. A yellow-vested cyclist on a white bike turns right.''

\section{Results}
\label{sec:main_results}
\subsection{Evaluation Strategy}
We evaluate eleven VLMs spanning four families (Qwen, InternVL, GPT, and Gemma) on \texttt{CrossView}. All results are obtained via uniformly sampling the VLM where frames are drawn independently from each camera. We report accuracy on multiple-choice question types (counting, temporal reasoning, event ordering, spatial reasoning, and camera identification) and evaluate open-ended summarization separately via ROUGE scores. Qwen, GPT, and Gemma models sample 16 frames per camera; InternVL models sample 8 frames per camera on the nuScenes and Ego-Exo4D subsets, and 4 frames on MEVA subset due to context-length constraints.

\begin{table*}[t]
    \centering
    \caption{Uniform sampling accuracy (\%) on the vehicle-centric and robot-centric subsets of \texttt{CrossView}. Results are reported for nuScenes and AgiBot across counting, event ordering, spatial, and temporal reasoning tasks.}
    \label{tab:robot}
    \resizebox{\textwidth}{!}{%
        \begin{tabular}{ll cccc cc}
            \toprule
            & & \multicolumn{4}{c}{nuScenes}
              & \multicolumn{2}{c}{AgiBot} \\
            \cmidrule(lr){3-6}\cmidrule(lr){7-8}
            Family & Model
              & Counting & Event Ordering & Spatial & Temporal
              & Temporal & Event Ordering \\
            \midrule
            \multirow{4}{*}{Qwen} & Qwen2.5-3B & 32.5 & 36.8 & 30.1 & 32.2 & 52.0 & 50.8 \\
             & Qwen2.5-7B & 27.5 & 24.4 & \textbf{45.9} & 21.6 & 58.8 & 54.4 \\
             & Qwen3-4B & 46.6 & \textbf{46.4} & 28.9 & 33.8 & 66.4 & 49.2 \\
             & Qwen3-8B & 43.2 & 41.2 & 33.1 & 36.0 & \textbf{69.6} & 49.2 \\
            \midrule
            \multirow{4}{*}{InternVL} & InternVL2-8B & 28.8 & 36.8 & 37.3 & 29.2 & 42.0 & 40.4 \\
             & InternVL2.5-8B & 37.1 & 28.4 & 32.1 & 33.4 & 60.0 & 46.4 \\
             & InternVL3.5-4B & 40.2 & 30.0 & 26.1 & 38.4 & 52.8 & 42.0 \\
             & InternVL3.5-14B & 46.2 & 36.4 & 43.3 & \textbf{44.8} & 54.8 & 47.6 \\
            \midrule
            \multirow{1}{*}{GPT} & GPT-5.2 & \textbf{47.8} & 29.2 & 36.5 & 25.4 & 66.8 & \textbf{66.0} \\
            \midrule
            \multirow{2}{*}{Gemma} & Gemma-3-4B & 44.7 & 41.6 & 18.6 & 38.2 & 50.8 & 36.8 \\
             & Gemma-3-12B & 43.8 & 43.2 & 26.4 & 35.6 & 60.4 & 41.2 \\
            \bottomrule
        \end{tabular}%
    }
\end{table*}

\begin{table*}[t]
    \centering
    \caption{Uniform sampling accuracy (\%) on the human-centric subsets of \texttt{CrossView}. Results are shown for Ego-Exo4D and MEVA across temporal reasoning, event ordering, counting, spatial reasoning, and best-camera identification tasks.}
    \label{tab:human}
    \resizebox{\textwidth}{!}{%
        \begin{tabular}{ll ccc ccccc}
            \toprule
            & & \multicolumn{3}{c}{Ego-Exo4D}
              & \multicolumn{5}{c}{MEVA} \\
            \cmidrule(lr){3-5}\cmidrule(lr){6-10}
            Family & Model
              & Temporal & Event Ordering & Best Camera
              & Counting & Event Ordering & Spatial & Temporal & Best Camera \\
            \midrule
            \multirow{4}{*}{Qwen} & Qwen2.5-3B & 50.0 & 48.0 & 20.6 & 22.9 & 21.9 & 41.8 & 32.5 & 27.3 \\
             & Qwen2.5-7B & 45.2 & 49.2 & 25.8 & 11.1 & \textbf{83.2} & 17.6 & 49.2 & 34.1 \\
             & Qwen3-4B & 48.8 & 50.0 & 33.2 & 20.4 & 34.7 & \textbf{49.0} & 48.9 & 34.8 \\
             & Qwen3-8B & 51.6 & 46.4 & 28.8 & 23.8 & 36.7 & 40.4 & 52.1 & \textbf{38.6} \\
            \midrule
            \multirow{4}{*}{InternVL} & InternVL2-8B & 28.8 & 43.6 & 20.2 & 18.9 & 41.4 & 47.3 & 52.8 & 36.6 \\
             & InternVL2.5-8B & 54.0 & 41.6 & 26.0 & 16.2 & 41.4 & 35.0 & 52.5 & 30.3 \\
             & InternVL3.5-4B & 42.8 & 41.6 & 23.8 & 14.3 & 27.9 & 46.5 & 51.9 & 36.1 \\
             & InternVL3.5-14B & \textbf{54.4} & 43.2 & 27.6 & 15.5 & 26.8 & 45.5 & \textbf{60.0} & 29.4 \\
            \midrule
            \multirow{1}{*}{GPT} & GPT-5.2 & 49.2 & \textbf{52.8} & \textbf{34.6} & 27.8 & 21.9 & 31.6 & 23.3 & 34.8 \\
            \midrule
            \multirow{2}{*}{Gemma} & Gemma-3-4B & 39.6 & 40.8 & 26.6 & \textbf{47.8} & 7.4 & 32.5 & 46.2 & 27.3 \\
             & Gemma-3-12B & 47.6 & 36.8 & 31.4 & 36.9 & 19.2 & 46.2 & 51.1 & 37.0 \\
            \bottomrule
        \end{tabular}%
    }
\end{table*}

\subsection{Main Results}
\begin{figure}[t]
    \centering
    \includegraphics[width=\linewidth]{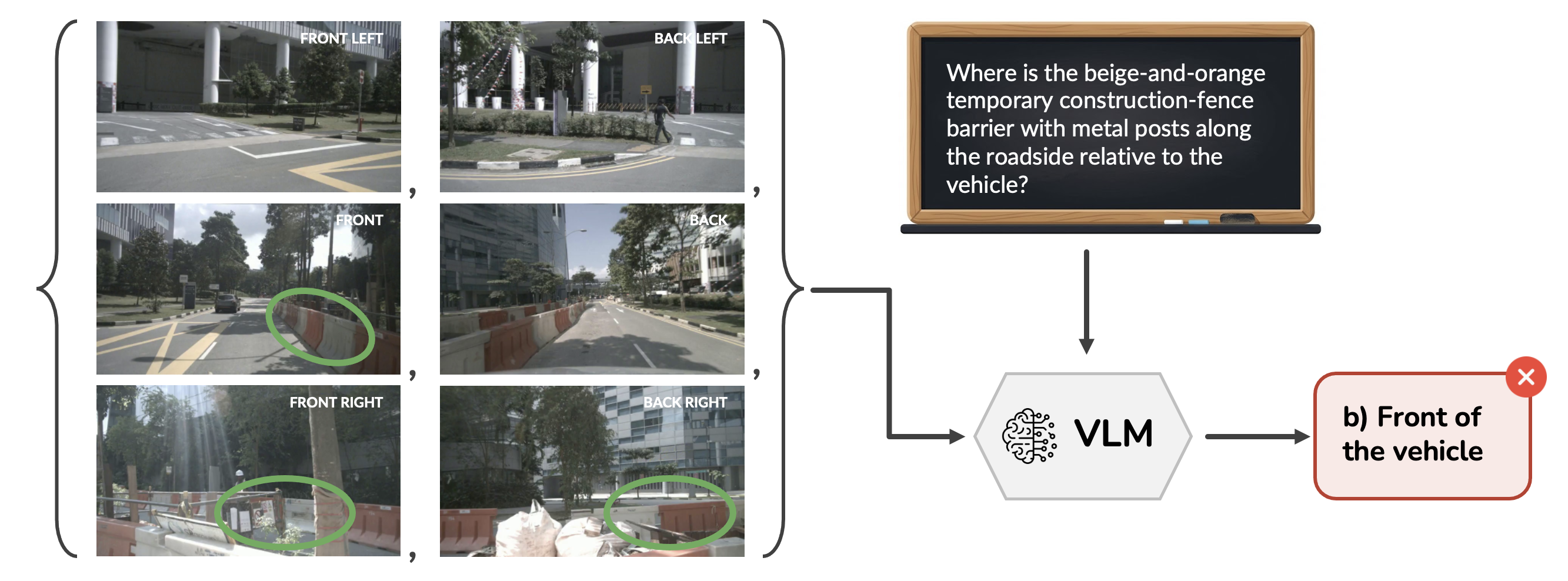}%
    \caption{\textbf{Qualitative failure case.} A driving example from \texttt{CrossView} under \underline{uniform sampling}, with frames drawn independently from each of the six surround-view cameras. The VLM localizes the roadside construction-fence barrier as \textbf{b)} \emph{Front of the vehicle}, likely because the front-right and back-right views capture it from a near front-on perspective; the correct answer is \textbf{c)} \emph{Right of the vehicle}.}
    \label{fig:qualitative_example}
\end{figure}

Multiple-choice accuracy results are reported in Tables~\ref{tab:robot} and~\ref{tab:human} for vehicle/robot-centric and human-centric benchmarks respectively. Summarization ROUGE scores are reported in Table~\ref{tab:summarization}. From our evaluations, we have the following insights:

\begin{enumerate}

\item \textbf{Multi-camera reasoning exposes a fundamental gap that model scale does not close.} Across all four benchmarks, no model family achieves consistently strong performance: notably, GPT-5.2 scores below 50\% on nuScenes temporal reasoning and below 35\% on camera identification in Ego-Exo4D, barely above the 25\% expected from random guessing on these questions (Tables~\ref{tab:robot} and~\ref{tab:human}). Crucially, these same models approach saturation on standard single-camera benchmarks, ruling out general visual or language capacity as the bottleneck. The deficit is uniform across architectures and parameter counts, pointing instead to a systematic absence of multi-camera understanding in current pretraining data. A representative failure is shown in Figure~\ref{fig:qualitative_example}, where the model collapses a right-side object into a front-of-vehicle judgment because individual camera views are reasoned over in isolation rather than jointly. Restricting inputs to a single camera confirms that many questions genuinely require evidence from multiple viewpoints (Table~\ref{tab:best_camera}), emphasizing that multi-camera reasoning is an inherently distinct problem.

\item \textbf{Tasks that require synthesizing information \emph{across} cameras are consistently the hardest.} Counting objects over multiple cameras and identifying the best camera for a given action, the two categories that require reasoning \emph{jointly} across synchronized feeds, consistently yield the lowest scores. Counting on MEVA and nuScenes falls within the 25–47\% range (Table~\ref{tab:robot}), and best-camera identification on Ego-Exo4D spans only 20–35\% across all models (Table~\ref{tab:human}), well below the 40–55\% those same models achieve on temporal and event-ordering tasks from identical videos, suggesting that these models are often guessing rather than reasoning about viewpoint utility.

\item \textbf{Scene scale and camera density are the primary drivers of benchmark difficulty.} Performance degrades monotonically as scene complexity grows. AgiBot, a dataset that involves a small, controlled environment with fixed-camera robot-manipulation tasks, yields the highest absolute accuracies (up to 69.6\% on temporal reasoning) whereas MEVA, a wide-area outdoor deployment with many overlapping, concurrently active cameras, produces the lowest scores in every category, most severely on counting and summarization tasks (Tables~\ref{tab:human} and~\ref{tab:summarization}). This showcases how complex scene and environment reasoning cannot be decomposed into a single-camera reasoning task.
\end{enumerate}

\begin{table*}[t]
    \centering
    \caption{Summarization performance on \texttt{CrossView} via ROUGE scores ($\times100$) under uniform frame sampling. R-1 and R-2 denote ROUGE-N overlap of unigrams and bigrams, respectively, and R-L denotes ROUGE-L, based on the longest common subsequence. Frame counts match the evaluation settings used in Tables~\ref{tab:robot} and~\ref{tab:human}.}
    \label{tab:summarization}
    \resizebox{\textwidth}{!}{%
        \begin{tabular}{ll cccc cccc cccc}
            \toprule
            & & \multicolumn{4}{c}{R-1}
              & \multicolumn{4}{c}{R-2}
              & \multicolumn{4}{c}{R-L} \\
            \cmidrule(lr){3-6}\cmidrule(lr){7-10}\cmidrule(lr){11-14}
            Family & Model
              & nuScenes & Ego-Exo4D & AgiBot & MEVA
              & nuScenes & Ego-Exo4D & AgiBot & MEVA
              & nuScenes & Ego-Exo4D & AgiBot & MEVA \\
            \midrule
            \multirow{4}{*}{Qwen} & Qwen2.5-3B & \textbf{26.3} & 28.8 & 34.4 & 9.5 & \textbf{5.3} & 4.3 & 7.9 & 0.5 & 15.0 & 17.7 & 21.8 & 7.6 \\
             & Qwen2.5-7B & 24.7 & 18.6 & 33.2 & \textbf{12.3} & 4.6 & 3.7 & \textbf{11.4} & 0.3 & 15.2 & 13.1 & \textbf{24.2} & \textbf{11.5} \\
             & Qwen3-4B & 17.9 & 25.9 & 21.8 & 6.7 & 4.9 & 4.3 & 6.3 & \textbf{1.0} & 11.8 & 15.5 & 14.3 & 4.9 \\
             & Qwen3-8B & 16.4 & 28.0 & 29.1 & 6.5 & 4.5 & 3.8 & 7.5 & 0.8 & 10.9 & 16.8 & 18.4 & 5.0 \\
            \midrule
            \multirow{4}{*}{InternVL} & InternVL2-8B & 16.1 & 23.0 & 31.4 & 5.7 & 4.0 & 2.6 & 6.2 & 0.6 & 10.8 & 15.8 & 21.2 & 4.2 \\
             & InternVL2.5-8B & 15.9 & 26.9 & 31.0 & 9.8 & 4.1 & 3.1 & 6.1 & 0.8 & 10.8 & 17.3 & 20.3 & 6.8 \\
             & InternVL3.5-4B & 23.7 & 29.3 & 31.9 & 6.2 & 5.2 & 4.2 & 6.6 & 0.5 & 15.4 & 18.3 & 20.6 & 4.9 \\
             & InternVL3.5-14B & 25.3 & 28.9 & 31.5 & 10.1 & 5.1 & 4.0 & 6.6 & 0.6 & \textbf{15.7} & 17.9 & 20.5 & 8.0 \\
            \midrule
            \multirow{1}{*}{GPT} & GPT-5.2 & 11.5 & \textbf{37.5} & \textbf{35.1} & 10.1 & 2.5 & \textbf{7.2} & 9.6 & 0.6 & 7.0 & \textbf{21.2} & 22.3 & 6.3 \\
            \midrule
            \multirow{2}{*}{Gemma} & Gemma-3-4B & 14.9 & 23.7 & 31.6 & 7.0 & 3.2 & 2.5 & 6.5 & 0.5 & 9.2 & 15.5 & 20.7 & 5.5 \\
             & Gemma-3-12B & 15.0 & 27.8 & 34.3 & 6.7 & 4.0 & 4.2 & 8.1 & 0.6 & 9.6 & 17.7 & 22.9 & 5.6 \\
            \bottomrule
        \end{tabular}%
    }
\end{table*}

\section{Discussion}
\subsection{Is There a Best Camera?}

A natural question is whether a single ``best'' camera can substitute for the full camera set and yield a similar accuracy. To investigate this, we compare feeding only the best single camera to Qwen2.5-7B-Instruct against the full multi-camera uniform results from Tables~\ref{tab:robot} and~\ref{tab:human}, shown in Table~\ref{tab:best_camera}. For Ego-Exo4D we test two candidates chosen for maximal scene coverage: the best-performing ground-truth exocentric camera (Best Exo) and the egocentric camera (Ego); for nuScenes, we select the front-facing vehicle camera (Front). 

We observe that no single camera reliably substitutes for the full camera set. Multi-camera inputs yield consistent gains on counting and spatial reasoning, tasks that inherently require integrating evidence across viewpoints, with the front-camera baseline on nuScenes dropping 8.8\% on counting and 4.4\% on spatial reasoning. Temporal reasoning on Ego-Exo4D similarly favors multi-camera input by roughly 4.5–4.8\% regardless of whether the egocentric or best exocentric view is used, indicating that even the most informative single perspective loses temporally salient cues captured by peripheral cameras.

The one reversal (nuScenes event ordering where the front camera leads by 7.6\%) \underline{does not} imply that a single view is superior. In reality, it is consistent with the finding in Section~\ref{sec:main_results} that models often struggle to synthesize evidence \emph{across} cameras: faced with redundant or conflicting views from six feeds, restricting input to the most semantically rich view can reduce noise on tasks with a dominant egocentric axis. Together, these results locate the bottleneck in the absence of principled cross-camera reasoning, a deficit that better camera selection alone cannot overcome.

\begin{table}[t]
    \centering
    \caption{Effect of restricting inputs to a single camera on \texttt{CrossView} for Qwen2.5-7B-Instruct. Compared to multi-camera inputs, single-camera views often reduce performance, showing that many questions require evidence from multiple viewpoints. Deltas are relative to the multi-camera baseline for Qwen2.5-7B from Tables~\ref{tab:robot} and~\ref{tab:human} (\textcolor{teal}{green} = single-camera better, \textcolor{red}{red} = multi-camera better, \textcolor{gray}{gray} = no change).}
    \label{tab:best_camera}
    \resizebox{0.9\columnwidth}{!}{%
    \begin{tabular}{llcccc}
        \toprule
        \textbf{Dataset} & \textbf{Input} & \textbf{Counting} & \textbf{Temporal} & \textbf{Event Ordering} & \textbf{Spatial} \\
        \midrule
        \multirow{2}{*}{Ego-Exo4D}
            & Best Exo & -- & 40.7 \textcolor{red}{($-$4.5)} & 47.3 \textcolor{red}{($-$1.9)} & -- \\
            & Ego & -- & 40.4 \textcolor{red}{($-$4.8)} & 48.8 \textcolor{red}{($-$0.4)} & -- \\
            \midrule
        nuScenes
            & Front & 18.7 \textcolor{red}{($-$8.8)} & 21.6 \textcolor{gray}{($\pm$0.0)} & 32.0 \textcolor{teal}{($+$7.6)} & 41.5 \textcolor{red}{($-$4.4)} \\
        \bottomrule
        \end{tabular}%
    }
\end{table}

\subsection{Uniform vs.\ Stitched Sampling}

We compare two multi-camera frame sampling strategies: \emph{uniform}, which samples $N$ frames independently per camera, and \emph{stitched}, which tiles all cameras into a single composite image per timestep (as seen in Figure~\ref{fig:teaser}). Table~\ref{tab:stitched_ablation} reports stitched results.

\begin{table}[t]
    \centering
    \caption{Comparison of uniform and stitched frame sampling on \texttt{CrossView} for Qwen2.5-7B. Stitched sampling improves performance on several tasks requiring cross-view spatial reasoning (\textit{e.g.}, counting and event ordering on nuScenes), but gains are dataset-dependent. Deltas are relative to the uniform baseline from Tables~\ref{tab:robot} and~\ref{tab:human} (\textcolor{teal}{green} = stitched better, \textcolor{red}{red} = uniform better).}
    \label{tab:stitched_ablation}
    \resizebox{0.9\columnwidth}{!}{%
        \begin{tabular}{llcccc}
            \toprule
            \textbf{Dataset} & \textbf{Strategy} & \textbf{Counting} & \textbf{Temporal} & \textbf{Event Ordering} & \textbf{Spatial} \\
            \midrule
            \multirow{2}{*}{nuScenes}
                & Uniform  & 27.5 & 21.6 & 24.4 & 45.9 \\
                & Stitched & 38.0 \textcolor{teal}{($+$10.5)} & 26.2 \textcolor{teal}{($+$4.6)} & 39.6 \textcolor{teal}{($+$15.2)} & 46.3 \textcolor{teal}{($+$0.4)} \\
            \midrule
            \multirow{2}{*}{AgiBot}
                & Uniform  & -- & 58.8 & 54.4 & -- \\
                & Stitched & -- & 58.4 \textcolor{red}{($-$0.4)} & 48.4 \textcolor{red}{($-$6.0)} & -- \\
            \midrule
            \multirow{2}{*}{Ego-Exo4D}
                & Uniform  & -- & 45.2 & 49.2 & -- \\
                & Stitched & -- & 53.6 \textcolor{teal}{($+$8.4)} & 48.0 \textcolor{red}{($-$1.2)} & -- \\
            \midrule
            \multirow{2}{*}{MEVA}
                & Uniform  & 11.1 & 49.2 & 83.2 & 17.6 \\
                & Stitched & 13.3 \textcolor{teal}{($+$2.2)} & 52.2 \textcolor{teal}{($+$3.0)} & 70.9 \textcolor{red}{($-$12.3)} & 34.2 \textcolor{teal}{($+$16.6)} \\
            \bottomrule
        \end{tabular}%
    }
\end{table}

Stitched sampling helps most on tasks that benefit from simultaneous spatial comparison across views. On nuScenes, where six cameras form a near-contiguous $360^{\circ}$ surround, stitching improves results across all categories, with counting and event ordering gaining 10.5 and 15.2 points respectively. The largest single gain occurs on MEVA spatial reasoning ($+16.6$), where wide-area scenes with many concurrent cameras benefit most from preserving inter-camera layout within a single context window. However, on AgiBot with the most controlled, low-clutter environment, stitching slightly degrades both metrics, suggesting that for simple scenes with few distractors, the added visual complexity of tiled frames introduces noise rather than useful context. Together, these results showcase how frame sampling is largely task-dependent, motivating work on adaptive frame sampling for complex scenes.

\section{Conclusion}
We introduced \texttt{CrossView}, a benchmark for evaluating multi-camera reasoning in VLMs across autonomous driving, surveillance, egocentric–exocentric interaction, and robotics. Unlike traditional VQA benchmarks that operate on a single visual stream, \texttt{CrossView} requires integrating information across synchronized viewpoints to answer questions involving counting, spatial reasoning, event ordering, and camera selection. Our evaluation reveals a consistent \textit{multi-camera reasoning gap}: even state-of-the-art VLMs struggle to combine evidence across views, with particularly low performance on tasks such as cross-view counting and identifying the most informative camera. This deficit persists across model families and scales, suggesting a structural limitation in how current models process and learn from multi-view data.

Real-world perception, from autonomous vehicles to robotic platforms, relies on \textit{networks} of cameras rather than single viewpoints, and \texttt{CrossView} exposes how far current models remain from reasoning effectively in these settings. We hope \texttt{CrossView} encourages the development of architectures, training objectives, and data pipelines that explicitly support cross-view evidence integration, advancing VLMs toward robust multi-camera understanding.

\section*{Acknowledgements}

This material is based upon work supported in part by the Office of Naval Research (ONR) under Grant No. N00014-22-1-2254. Additionally, this work was supported by the Defense Advanced Research Projects Agency (DARPA) contract DARPA ANSR: RTX CW2231110. Approved for Public Release, Distribution Unlimited.


%
%
\bibliographystyle{splncs04}
\bibliography{main}

\clearpage
\appendix
\section{Prompts for Data Generation}

Question generation in \texttt{CrossView} proceeds in two stages: (1) a \emph{structured extraction} stage, where per-dataset annotations (scene graphs, activity logs, or camera metadata) are parsed into typed records, and (2) a \emph{verbalization} stage, where an LLM is prompted to convert those records into natural-language questions, multiple-choice options, and ground-truth answers.
Each prompt below corresponds to one question category and one dataset context:

\begin{itemize}
  \item \textbf{Counting}. Applied after object-instance counts are aggregated across all frames of a scene. The LLM selects a salient object class and forms a counting question with an exact numeric answer.
  \item \textbf{Event Ordering}. Applied after a chronologically-sorted list of distinct events is extracted from the annotation log. The LLM scrambles the events, assigns roman-numeral labels, and produces a re-ordering multiple-choice question (MCQ).
  \item \textbf{Spatial}. Applied after the scene graph for a single representative frame is filtered to directional relationships (\emph{in front of}, \emph{behind}, \emph{left/right of}). The LLM selects an unambiguous object pair and forms a spatial-relationship MCQ.
  \item \textbf{Temporal}. Applied after a grounding event and a target event are identified together with their temporal relationship (\emph{before}, \emph{after}, \emph{during}, \emph{in-between}). The LLM produces a before/after/simultaneous MCQ with visually-grounded distractors.
  \item \textbf{Summarization}. Applied after a condensed temporal scene graph is assembled from per-frame captions. The LLM produces a 2--3 sentence reference summary used as the ground-truth open-ended answer.
  \item \textbf{Best Camera}. Applied after the best exocentric camera for a scene or event is identified from annotation metadata. The LLM generates a scene-description question asking which exocentric view best captures the depicted activity.
\end{itemize}

\begin{tcolorbox}[breakable,colback=gray!5!white,colframe=black!75!black,title=Counting Question Generator Prompt Prompt,fonttitle=\bfseries]
\scriptsize
\begin{Verbatim}[breaklines=true]
You are an expert at creating video understanding questions for autonomous driving datasets. Given object counts from a nuScenes scene, generate ONE natural, specific counting question with its answer.

TASK: Generate a counting question about the number of instances of an object type in the video.

GUIDELINES:
1. Ask about the total count of a specific object class across the entire video
2. Make the question natural and conversational
3. Focus on interesting or varied object classes (not always cars)
4. Include the exact count as the answer
5. Format: Return a JSON object with "question" and "answer" fields

EXAMPLES:

Example 1:
Object counts:
- vehicle.car: 15
- human.pedestrian.adult: 3
- vehicle.bicycle: 2

Generated:
{{
  "question": "How many pedestrians appear in this driving scene?",
  "answer": "3",
  "reasoning": "There are 3 unique human.pedestrian.adult instances across all frames in the video."
}}

Example 2:
Object counts:
- vehicle.car: 28
- vehicle.truck: 5
- human.pedestrian.child: 1
- vehicle.bicycle: 4

Generated:
{{
  "question": "What is the total number of bicycles visible throughout the video?",
  "answer": "4",
  "reasoning": "There are 4 unique vehicle.bicycle instances tracked across the scene."
}}

Example 3:
Object counts:
- vehicle.car: 12
- vehicle.bus: 1
- vehicle.motorcycle: 2
- human.pedestrian.adult: 7

Generated:
{{
  "question": "How many motorcycles can be seen in the entire video sequence?",
  "answer": "2",
  "reasoning": "The scene contains 2 unique vehicle.motorcycle instances."
}}

NOW GENERATE A QUESTION FOR THIS SCENE:

Scene: {scene_token}
Number of frames: {num_frames}
Object counts:
{object_counts}

Generate ONE counting question following the format above. Choose an interesting object class that has a meaningful count (avoid extremely common objects like parked cars unless the count is notable). Return only the JSON object.
\end{Verbatim}
\end{tcolorbox}

\begin{tcolorbox}[breakable,colback=gray!5!white,colframe=black!75!black,title=Event Ordering Generator Prompt,fonttitle=\bfseries]
\scriptsize
\begin{Verbatim}[breaklines=true]
You are an expert at creating temporal logic questions for autonomous driving datasets. You will be given a list of distinct events and the description of the object (in parenthesis) in the event occurring in a specific chronological order (from start to finish). Your job is to create a "re-ordering" multiple choice question.

TASK: Generate ONE event ordering multiple choice question. You must first randomize the presentation of the events in the question text, adjust the events with the object descriptions for proper grounding, and ask the agent to identify the correct chronological sequence.

GUIDELINES:
1. **Input Data:** You are provided with a list of events: Event 1 -> Event 2 -> Event 3 -> Event 4. This is the GROUND TRUTH order.
2. **Question Construction:**
   - List the events in the question body in a **RANDOM/SCRAMBLED** order.
   - **Crucial:** You must incorporate the object descriptions (from the parentheses) into the event text within the question. Do not ignore the visual details.
   - Assign them temporary labels (e.g., I, II, III, IV or A, B, C, D) inside the question text.
   - Ask "Which is the correct chronological order of these events?"
3. **Options:**
   - The correct option must correspond to the Ground Truth sequence (e.g., "II -> IV -> I -> III").
   - The three distractors must be incorrect permutations (e.g., reverse order, or swapped middle steps).
   - The position of the correct answer should be randomized.
4. **Clarity:** Ensure the event descriptions are distinct enough that a human observer could logically deduce the order.
5. **Format:** Return a JSON object with "question", "options", "answer", and "reasoning".

EXAMPLES:

Example 1: 3 Events
Ground Truth Input: 
1. The traffic light turns red (the traffic light is on the grey pole).
2. The SUV comes to a stop (the SUV is white in color).
3. Pedestrians enter the crosswalk (the pedestrian is wearing a white tshirt and appears to be crossing the road).

Output:
{{
  "question": "Identify the correct chronological order of the following related events observed in the scene:\n(A) A pedestrian wearing a white tshirt enters the crosswalk\n(B) The traffic light on the grey pole turns red\n(C) The white SUV comes to a stop",
  "options": {{
    "A": "B -> C -> A",
    "B": "A -> B -> C",
    "C": "C -> A -> B",
    "D": "B -> A -> C"
  }},
  "answer": "A",
  "reasoning": "The logical traffic sequence is: Light turns red (B) -> Vehicle stops (C) -> Pedestrians feel safe to enter (A)."
}}

Example 2: 4 Events
Ground Truth Input:
1. A delivery truck activates its turn signal (delivery truck is white and is an 18 wheeler).
2. The truck merges into the right lane (delivery truck is white and is an 18 wheeler. The road has 3 lanes).
3. The truck slows down approaching the driveway (delivery truck is white and is an 18 wheeler).
4. The truck turns into the delivery bay (delivery truck is white and is an 18 wheeler, the delivery bay appears to be in a gray building with a large hangar).

Output:
{{
  "question": "The following four events occurred involving a white 18-wheeler delivery truck. Arrange them in the order they happened:\nI. The truck turns into the delivery bay of a gray building with a large hangar\nII. The truck merges into the right lane of the 3-lane road\nIII. The truck activates its turn signal\nIV. The truck slows down approaching the driveway",
  "options": {{
    "A": "III -> II -> IV -> I",
    "B": "II -> III -> I -> IV",
    "C": "I -> IV -> II -> III",
    "D": "III -> IV -> II -> I"
  }},
  "answer": "A",
  "reasoning": "Standard driving maneuver sequence: Signal (III) -> Merge (II) -> Decelerate (IV) -> Turn (I)."
}}

Example 3: 4 Events
Ground Truth Input:
1. A delivery truck activates its turn signal (the truck is blue with a white logo on the side).
2. Cyclist on the right side of the road stops (the cyclist is wearing a red helmet and riding a silver bike).
3. The truck turns into a building (the building has a large glass entrance).
4. Construction worker enters a building (the worker is wearing a high-vis orange vest and hard hat).

Output:
{{
  "question": "The following four events occurred in the scene. Arrange them in the order they happened:\nI. A cyclist wearing a red helmet on a silver bike stops on the right side\nII. A construction worker wearing a high-vis orange vest enters a building\nIII. A blue delivery truck with a white logo activates its turn signal\nIV. The truck turns into a building with a large glass entrance",
  "options": {{
    "A": "III -> II -> IV -> I",
    "B": "II -> III -> I -> IV",
    "C": "I -> IV -> II -> III",
    "D": "III -> I -> IV -> II"
  }},
  "answer": "D",
  "reasoning": "The video shows the truck signaling (III), the cyclist stopping (I), the truck turning (IV), and finally the worker entering (II)."
}}

Example 4: 3 Events
Ground Truth Input:
1. A delivery truck activates its turn signal (the truck is a small red van).
2. The truck merges into the right lane (the lane markings are dashed white lines).
3. The truck slows down approaching the driveway (the driveway leads to a brick residential house).

Output:
{{
  "question": "The following three events occurred involving a small red van. Arrange them in the order they happened:\nI. The van merges into the right lane with dashed white lines\nII. The van activates its turn signal\nIII. The van slows down approaching the driveway of a brick residential house",
  "options": {{
    "A": "II -> III -> I",
    "B": "I -> II -> III",
    "C": "III -> I -> II",
    "D": "II -> I -> III"
  }},
  "answer": "D",
  "reasoning": "Standard driving maneuver sequence as shown in the video: Signal -> Merge -> Slow down."
}}

NOW GENERATE A QUESTION FOR THIS SCENE:

Ground Truth Events (Chronological):
{events_list}

Instructions: Scramble the events in the question text. Incorporate the visual descriptions from the parentheses into the question text. Provide 4 option sequences. Return strictly valid JSON.
\end{Verbatim}
\end{tcolorbox}

\begin{tcolorbox}[breakable,colback=gray!5!white,colframe=black!75!black,title=Spatial Question Generator Prompt,fonttitle=\bfseries]
\scriptsize
\begin{Verbatim}[breaklines=true]
You are an expert at creating MCQ spatial reasoning questions from videos with multiple camera scenes. Given a scene description and scene graph relationships, generate ONE multiple-choice question with four options about a directional spatial relationship between two objects.

TASK: Create a single MCQ about a directional relationship using ONLY these relations: "in front of", "behind", "to the left of", "to the right of".

GUIDELINES:
1. Use the scene description to ground the question (brief, 1 sentence max).
2. Ask about the relative position between two specific object types or instances.
3. Enforce a unique correct answer:
   - Prefer pairs where the source object class has a single instance in the scene/frame.
   - If not possible, use a descriptive reference (e.g., "the red bus", "the parked white car") when available in the description or graph.
   - Ensure none of the distractor options is also true in the scene.
4. Avoid vague relations like "near" or "next to". Only use the four directional relations above.
5. If no unique answer can be guaranteed for the sampled pair, choose another pair.
6. Output format MUST be a JSON object with fields: question, options (A-D), correct_option, rationale.

INPUTS:
- Scene token: {scene_token}
- Frame index: {frame_idx}
- Scene description (short): {scene_description}
- Candidate directional relationships (filtered):
{directional_relationships}
- Object instance counts (by class):
{object_counts}

EXAMPLES:

Example A:
Scene description: A cyclist passes a stopped car at an intersection.
Relationships:
- Frame 12: vehicle.bicycle is to the right of vehicle.car
- Frame 12: vehicle.car is behind vehicle.truck

Generated:
{
  "question": "In this scene [scene description], where is the bicycle relative to the car?",
  "options": {
    "A": "In front of the car",
    "B": "Behind the car",
    "C": "To the left of the car",
    "D": "To the right of the car"
  },
  "correct_option": "D",
  "rationale": "At frame 12, the scene graph shows the bicycle is to the right of the car."
}

Example B:
Scene description: A bus approaches with a single pedestrian waiting at the curb.
Relationships:
- Frame 8: human.pedestrian.adult is in front of vehicle.bus

Generated:
{
  "question": "In the scene description, what is directly in front of the bus?",
  "options": {
    "A": "Car",
    "B": "Pedestrian",
    "C": "Bicycle",
    "D": "Wheelchair"
  },
  "correct_option": "B",
  "rationale": "At frame 8, the pedestrian is in front of the bus according to the scene graph."
}

Example C:
Scene description: A bus approaches with a single pedestrian waiting at the curb.
Relationships:
- Frame 8: human.pedestrian.adult is in front of vehicle.bus

Generated:
{
  "question": "Which of the statements is true for the scene?",
  "options": {
    "A": "A bus is in front of the pedestrian",
    "B": "A pedestrian is in front of the bus",
    "C": "A bicycle is to the right of the bus",
    "D": "A wheelchair is in front of the pedestrian"
  },
  "correct_option": "B",
  "rationale": "At frame 8, the pedestrian is in front of the bus according to the scene graph."
}

NOW GENERATE ONE MCQ:
- Use exactly one relationship from the input that yields a unique answer.
- If multiple are valid, choose the one with the fewest instances of the source object class.
- Return ONLY the JSON object described in the schema.
\end{Verbatim}
\end{tcolorbox}

\begin{tcolorbox}[breakable,colback=gray!5!white,colframe=black!75!black,title=Temporal Reasoning Prompt,fonttitle=\bfseries]
\scriptsize
\begin{Verbatim}[breaklines=true]
You are an expert at creating video understanding questions for autonomous driving datasets. You will be given a list of objects and their descriptions and activity in the temporal order they appear in the video. Your job is to create temporal reasoning questions from this list.

TASK: Generate ONE temporal reasoning multiple choice question with 4 options and its answer pair based on the provided list of objects and their activities over time.

GUIDELINES:
1. **Input Data:** You will be provided with:
   - **Grounding Object/Event:** The reference event with a visual description (e.g., "Car stops (car is red)").
   - **Target Object/Event:** The event to be identified, also with a visual description (e.g., "Pedestrian crosses (wearing a blue hat)").
   - **Temporal Relationship:** The logic to apply (Before, After, In-between, During).
2. **Visual Specificity:** You **MUST** incorporate the specific visual descriptions (color, vehicle type, clothing, location) provided in the input for **BOTH** the Grounding object and the Target object. 
   - The **Question** should usually describe the Grounding object visually to set the context.
   - The **Correct Answer** should usually describe the Target object visually to confirm identification.
3. **Question Types:**
   - **Before/After:** "What did the [Target + desc] do immediately after [Grounding + desc] happened?"
   - **In-Between:** "Between [Grounding Start + desc] and [Grounding End + desc], what did the [Target + desc] do?"
   - **Simultaneous:** "While the [Grounding + desc] was happening, what was the [Target + desc] doing?"
4. **Distractors:** Generate three plausible but incorrect distractors. They should describe realistic driving behaviors but be factually wrong based on the input.
5. **Format:** Return a JSON object with "question", "options", "answer", and "reasoning".

EXAMPLES:

Example 1: In-Between Logic
Grounding Input: 
- Start: A sedan turns right (the sedan is blue and has a sunroof).
- End: A city bus stops (the bus is red and displays route 55).
Target Input: A cyclist rides through traffic (the cyclist is wearing a high-vis yellow vest).
Temporal Relationship: In-between

Output:
{{
  "question": "What activity occurred in the time between the blue sedan with a sunroof initiating its right turn and the red route 55 bus coming to a stop?",
  "options": {{
    "A": "A pedestrian in a blue coat began crossing the street.",
    "B": "A cyclist wearing a high-vis yellow vest rode through the traffic.",
    "C": "The ego-vehicle performed a U-turn.",
    "D": "A green delivery truck moved into the bus lane."
  }},
  "answer": "B",
  "reasoning": "The target event (cyclist in yellow vest) happens chronologically between the two grounding events."
}}

Example 2: Before/After Logic
Grounding Input: A hatchback clears the intersection (the hatchback is white and covered in mud).
Target Input: A taxi merges into the left lane (the taxi is a yellow SUV).
Temporal Relationship: Immediately after

Output:
{{
  "question": "What did the yellow SUV taxi do right after the mud-covered white hatchback cleared the intersection?",
  "options": {{
    "A": "It merged into the left lane.",
    "B": "It came to a sudden stop.",
    "C": "It honked at a pedestrian.",
    "D": "It turned into a parking lot."
  }},
  "answer": "A",
  "reasoning": "The grounding event is the muddy hatchback clearing the path, immediately followed by the yellow SUV taxi's merge."
}}

Example 3: During/Simultaneous Logic
Grounding Input: The ego-vehicle waits for schoolchildren to cross (the children are wearing red backpacks).
Target Input: A van idles behind the ego-vehicle (the van is white with a floral delivery logo).
Temporal Relationship: During

Output:
{{
  "question": "While the ego-vehicle was waiting for the schoolchildren with red backpacks to finish crossing, what was the white van with the floral logo doing?",
  "options": {{
    "A": "It attempted to overtake the ego-vehicle.",
    "B": "It reversed to find another route.",
    "C": "It remained idling behind the ego-vehicle.",
    "D": "It turned on its hazard lights."
  }},
  "answer": "C",
  "reasoning": "The target state (idling white van with floral logo) occurs simultaneously with the grounding state (waiting for children with red backpacks)."
}}

NOW GENERATE A QUESTION FOR THIS SCENE:

Grounding Object/Event: {grounding_input}
Target Object/Event: {target_input}
Temporal Relationship: {rel_type}

Instructions: Return strictly valid JSON containing "question", "options", "answer", and "reasoning". Ensure the question and correct answer explicitly use the visual descriptions provided in the input parentheses.
\end{Verbatim}
\end{tcolorbox}

\vspace{1em}

\begin{tcolorbox}[breakable,colback=gray!5!white,colframe=black!75!black,title=Summarization Prompt,fonttitle=\bfseries]
\scriptsize
\begin{Verbatim}[breaklines=true]
You are an expert at summarizing driving scenes from multi-camera autonomous vehicle data. Given a temporal scene graph that describes objects, their activities, spatial relationships, and evolution across frames (from the ego-vehicle's perspective across all camera views), produce a concise summary.

TASK: Summarize the ego-actor's interactions across all views in 2-3 sentences.

QUESTION (fixed, for reference): "Provide a comprehensive summary of the ego-actor's interactions across all views."

GUIDELINES:
1. Focus on what the ego-vehicle (self-driving car) encountered and interacted with over the scene
2. Mention key objects (vehicles, pedestrians, cyclists, etc.), their activities, and salient spatial/temporal patterns
3. Condense the temporal scene graph into a coherent narrative - avoid listing frame-by-frame details
4. Write in natural, fluent language
5. Keep the response to 2-3 sentences maximum

INPUT - Temporal Scene Graph (condensed from scene graph + captions):
---
{temporal_scene_graph}
---

Generate the summary now. Return ONLY the 2-3 sentence summary, no preamble or labels.
\end{Verbatim}
\end{tcolorbox}

\vspace{1em}

\begin{tcolorbox}[breakable,colback=gray!5!white,colframe=black!75!black,title=Best Camera Prompt,fonttitle=\bfseries]
\scriptsize
\begin{Verbatim}[breaklines=true]
You are an expert at creating video understanding questions focusing on camera angles for ego/exo task videos. You will be given information about camera perspectives related to a scene or a specific event, along with a list of all available exocentric camera angles. Your job is to create a multiple-choice question with 4 options and its answer.

TASK: Generate ONE camera perspective multiple choice question with 4 options and its answer pair based on the provided information.

GUIDELINES:
1. The question should focus on identifying the best EXOCENTRIC (third-person) camera angle for a given scene description or a specific event within the scene.
2. The options should include the correct exocentric camera angle and three plausible but incorrect exocentric camera angles from the provided list of all available cameras.
3. IMPORTANT: If any description contains "C" (which stands for the camera-wearer or person), replace it with "the camera-wearer" or "the person" as appropriate for natural language. For instance, "C presses" should become "The camera-wearer presses" or "The person presses".
4. For "Best camera angle for the scene": the Description field contains a chronological list of video annotations. Derive a single-sentence summary of what the camera-wearer is doing and incorporate it into the question (e.g. "This video shows the camera-wearer <summary>. Which exocentric camera best captures the camera-wearer's actions?").
5. IMPORTANT: The question MUST always ask for the best EXOCENTRIC camera. Always use the phrase "exocentric camera" — never use vague phrasing like "which camera".

FORMAT: Return a JSON object with "question", "options", "answer", and "reasoning".

EXAMPLE 1: Best camera angle for the scene
Annotations:
- C picks up the bicycle wheel
- C inspects the tire for damage
- C removes the tire from the rim
- C installs a new inner tube
- C reassembles the wheel onto the bicycle
Best Camera for Scene: camera 4
All Available Cameras: camera 1, camera 2, camera 3, camera 4

Output:
{{
  "question": "This video shows the camera-wearer repairing a bicycle by removing, inspecting, and reassembling the wheel. Which exocentric camera best captures the camera-wearer's actions throughout this scene?",
  "options": {{
    "A": "camera 1",
    "B": "camera 2",
    "C": "camera 3",
    "D": "camera 4"
  }},
  "answer": "D",
  "reasoning": "Camera 4 provides the most comprehensive and informative exocentric view of the overall bicycle repair scene."
}}

EXAMPLE 2: Best camera angle for a specific event
Event Description: The person holds the bicycle wheel with both of his hands (timestamp=30.81681).
Best Camera for Event: camera 1
All Available Cameras: camera 1, camera 2, camera 3, camera 4, camera 5

Output:
{{
  "question": "Which exocentric camera best shows the person holding the bicycle wheel with both of his hands?",
  "options": {{
    "A": "camera 1",
    "B": "camera 2",
    "C": "camera 3",
    "D": "camera 4",
    "E": "camera 5"
  }},
  "answer": "A",
  "reasoning": "Camera 1 offers the clearest and most detailed exocentric perspective of the person's action of holding the bicycle wheel with both hands at 30.81681 seconds."
}}

EXAMPLE 3: Best camera angle for multiple events
Event 1: The person holds the bicycle wheel with both hands (timestamp=30.82); Event 2: The person inflates the bicycle tire (timestamp=95.43)
Best Camera for Events: camera 1
All Available Cameras: camera 1, camera 2, camera 3, camera 4, camera 5

Output:
{{
  "question": "Which exocentric camera best shows the person holding the bicycle wheel with both hands and then inflating the bicycle tire?",
  "options": {{
    "A": "camera 1",
    "B": "camera 2",
    "C": "camera 3",
    "D": "camera 4",
    "E": "camera 5"
  }},
  "answer": "A",
  "reasoning": "Camera 1 provides the best exocentric view of both events: the person holding the bicycle wheel at 30.82 seconds and then inflating the tire at 95.43 seconds."
}}

NOW GENERATE A QUESTION FOR THIS SCENE:

Question Type: {question_type}
Description (if applicable): {description}
Best Camera: {best_camera}
All Available Cameras: {all_cameras}

Instructions: Return strictly valid JSON containing "question", "options", "answer", and "reasoning".
\end{Verbatim}
\end{tcolorbox}

\end{document}